\documentclass[a4paper]{article}

\usepackage{INTERSPEECH_v2}
\usepackage{amssymb,epsfig,graphics,graphicx,tipa,times,amsmath,algorithmic,algorithm,multirow}

\newcommand{\argmin}{\arg\!\min}
\newcommand{\tabincell}[2]{\begin{tabular}{@{}#1@{}}#2\end{tabular}}

\title{Acoustic data-driven lexicon learning based on a greedy pronunciation selection framework}

\name{ Xiaohui Zhang$^{1}$, Vimal Manohar$^{1,2}$, Daniel Povey$^{1,2}$, Sanjeev Khudanpur$^{1,2}$
\thanks{
This work was partially supported by 
DARPA LORELEI Grant No HR0011-15-2-0024, NSF Grant No CRI-1513128 and IARPA Contract No 2012-12050800010.
The U.S. Government is authorized to reproduce and distribute reprints for Governmental
purposes notwithstanding any copyright annotation thereon. The views and
conclusions contained herein are those of the authors and should not be interpreted
as necessarily representing the official policies or endorsements,
either expressed or implied, of DARPA, IARPA, DoD/ARL or the U.S. Government. Thanks Dr. Jack Godfrey, Dr. Alan MaCree, Dr. Mengyang Gu, Chloe Haviland for useful discussions.
}
}

\address{$^{1}$Center for Language and Speech Processing\\
 $^{2}$Human Language Technology Center of Excellence\\
  The Johns Hopkins University, Baltimore, MD 21218, USA
  }
  \email{ xiaohui@jhu.edu, vimal.manohar91@gmail.com, danielpovey@gmail.com, khudanpur@jhu.edu}
\begin{document}

\maketitle
\begin{abstract}
Speech recognition systems for irregularly-spelled languages
like English normally require hand-written pronunciations.
In this paper, we describe a system for automatically 
obtaining pronunciations of words for which pronunciations
are not available, but for which transcribed
data exists.  Our method integrates information from the
letter sequence and from the acoustic evidence. The novel aspect of the problem that we address is the problem of how to prune entries
from such a lexicon (since, empirically, lexicons with too many
entries do not tend to be good for ASR performance).
Experiments on various ASR tasks show that, with the proposed framework, starting with an initial lexicon of several thousand words, we are able to learn a lexicon which performs close to a full expert lexicon in terms of WER performance on test data, and is better than lexicons built using G2P alone or with a pruning criterion based on pronunciation probability.
\end{abstract}
%
\noindent\textbf{Index Terms}: speech recognition, pronunciation lexicon learning
\section{Introduction}
In the past few years, there has been an growing interest in investigating acoustic data-driven lexicon learning for continuous speech recognition, i.e. automatically 
obtaining pronunciations of words for which pronunciations
are not available , but for which transcribed acoustic
data exists.  In order to develop ASR systems under limited lexicon resources, one solution is to adopt a graphemic lexicon~\cite{gales2015unicode,harwath2014speech} or acoustic unit discovery methods~\cite{lee2015unsupervised,lee2013joint}, which totally eliminate the expert efforts for developing a phonetic pronunciation lexicon. In real applications, however, a more common scenario is that we already have a phonetic inventory, and a small expert lexicon for a specific language. Our work focuses on this case, i.e. given a small expert lexicon, we want to derive pronunciations for Out-of-Vocabulary (OOV) words, for which we know the text form and have acoustic examples. 

Given a small expert lexicon, the most straightforward way to generate pronunciation candidates for OOV words is to train a Grapheme-to-Phoneme (G2P) \cite{bisani2008joint} model 
using the seed lexicon and apply it to these OOV words~\cite{lu2013acoustic,goel2010approaches,mcgraw2013learning}.  But for languages like English,
and for proper names and abbreviations,
G2P does not always give high quality
pronunciations.  Pronunciations from phonetic decoding
can help to fill this gap.
Previous work has combined these with G2P-generated pronunciations~\cite{rasipuram2012combining,laurent2010acoustics},  or added into G2P training examples ~\cite{goel2010approaches,chen2016acoustic,tsujioka2016unsupervised}.  In the work we describe here, we use candidates from both
G2P and phonetic decoding.

The aspect of the problem that we focus on is candidate pruning.  That is, given a set of pronunciation candidates from G2P and phonetic decoding (and maybe some from a manually created lexicon), which subset should we keep? Keeping
all the pronunciations is impractical because it would make decoding slow,
and also because too many pronunciations tend to hurt ASR performance, even when pronunciation probabilities are used~\cite{hain2002implicit}.  

Previous work on candidate pruning has relied on estimated pronunciation
probabilities to determine which candidates should be cut~\cite{chen2016acoustic,lu2013acoustic,mcgraw2013learning,goel2010approaches,tsujioka2016unsupervised}. The main defect with this is that for words with multiple pronunciations, it tends to give us too many minor pronunciation variants (e.g. reflecting co-articulation effects),  which is undesirable for ASR.  If we rely on pronunciation probabilities alone it is hard to discard 
those types of variants while keeping variants that come from different meanings of the word.  

The core idea of this paper is a likelihood-based criterion for
pronunciation-candidate pruning that naturally keeps candidates that are ``far apart''.  

This paper is organized as follows.  We discuss how we generate
pronunciation candidates in Section~\ref{phonetic}; we
explain how we collect acoustic evidence from training data
in Section~\ref{evidence}.  We explain our likelihood-based pronunciation selection strategy in Section~\ref{selection}.  Experimental results on various ASR tasks are provided in Section~\ref{exp}, and we conclude
in Section~\ref{sec:conclusions}.
\begin{figure} 
\centering
\includegraphics[width=0.45  \textwidth]{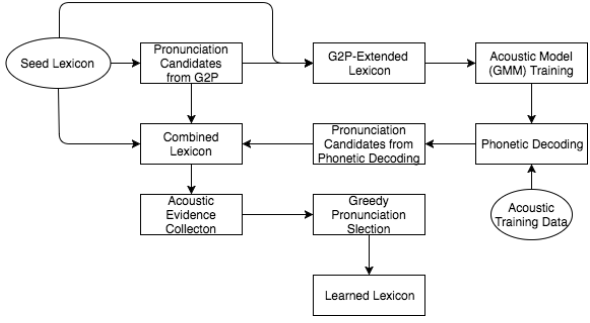}
\caption{\label{fig:diagram}The proposed framework of acoustic-data driven lexicon learning.}
\end{figure}
\section{Collecting pronunciation candidates from multiple sources}
\label{phonetic}
\vspace{-0.21em}
In our framework, like~\cite{laurent2010acoustics}, we first extend the seed lexicon to include OOV words in the training data, using a G2P model trained on the seed lexicon, and then train an acoustic model (AM) using the G2P-extended lexicon.  Then we generate alignments for all training data, based on which we then train a bi-gram phone language model (LM). Using this phone LM and the AM, we construct a phonetic decoder and use it to generate phonetic transcription of training data. For each individual word token in the transcript, we can align it with a phone sequence using timing information from the alignments and phonetic transcriptions. Then for each specific word $w$, we can compute the relative frequency of each phone sequence being aligned to it, by normalizing each phone sequence's count by the most frequent phone sequence's count. Then we filter out those phone sequences whose relative frequency is too low (e.g. smaller than 0.1) and keep the left ones as the alternative pronunciations generated from phonetic decoding. Then we combine these  alternative pronunciation candidates with the G2P-extended lexicon into a large lexicon (called combined lexicon). For each word $w$ from the combined lexicon, let $\mathcal{B}$ denote the set of pronunciation candidates collected from multiple sources, and $b$ denote one pronunciation (baseform) candidate. The source of $b$ (denoted as $s(b)$) could be one of the three: G2P/phonetic decoding. In the next section we will specify how we collect acoustic evidence for all pronunciation candidates in $\mathcal{B}$ .
\section{Acoustic evidence collection}
\label{evidence}
First we introduce some notations. Let $\textbf{O} = \{ \mathcal{O}_1, \mathcal{O}_2,...,\mathcal{O}_M \}$ denote acoustic sequences; $M_w$ denote the number of utterances in $\textbf{O}$ which contain the word $w$ \footnote{we assume that each word appears in each utterance's transcript at most once. In practice, if a word appears multiple times in an utterance, we divide the utterance into sub-utterances where each one only contains one token of the word.}; Then we further define $\theta_{wb} \triangleq p(w, b) $ as the pronunciation probability of a pronunciation $b$ for a word $w$ ($\sum_{b \in \mathcal{B}} \theta_{wb} = 1$), and $\boldsymbol{\theta}_w \triangleq \{\theta_{wb}: b \in \mathcal{B}\}$  as the pronunciation model for word $w$. 
We define $\tau_{uwb}  \triangleq p(\mathcal{O}_u | w, b)$ as the conditional data likelihood given the pronunciation of $w$ being $b$, which is determined by the acoustic model. This is the "acoustic evidence" we want to derive from lattice statistics, which is needed by our pronunciation selection algorithm.

With the combined lexicon and an existing  AM (the one we used for phonetic decoding in the candidate collection phase), we generate lattices for each training utterance. This lattice generation treats distinct pronunciations of words as distinct symbols for the purposes of lattice determinization, unlike our standard procedure described in~\cite{povey2012generating}. This is achieved by putting both phone symbols and word symbols as the input sequence on the FST prior to lattice determinization. From the lattices, we can obtain per-utterance lattice pronunciation-posterior statistics $\gamma_{uwb} \triangleq p(w, b | \mathcal{O}_u)$.  

When the lattices were generated, we assign uniform priors over all pronunciation candidates of each word in the combined lexicon. By Bayes' rule, we can directly use the posterior statistics $\gamma_{uwb}$ as the likelihoods $\tau_{uwb}$ \footnote{Strictly speaking, Bayes' rule only gives us $\tau_{uwb} \propto \gamma_{uwb}$, i.e. $\tau_{uwb}$ can only be treated as $\gamma_{uwb}$ up to a constant, but the constant doesn't affect the objective \eqref{eqn:objf} we want to optimize.}.
Because lattices are pruned, a posterior  $\gamma_{uwb}$ could be zero even if $w$ actually appears in a utterance $u$. So we always floor $\gamma_{uwb}$ to a small positive scalar $\delta$ (In practice it's set between $10^{-7}$ and $10^{-5}$), so that we have $\tau_{uwb} \geq \delta, \forall u, w, b$. 

Based on $\gamma_{uwb}$, we can obtain another useful statistic, the average pronunciation posterior $\gamma_{wb} \triangleq \frac{1}{M_w} \sum_{u} \gamma_{uwb}$, where the summation $\sum_{u}$ is only taken over those utterances where the word $w$ actually appears.

After the lattices were dumped, for each word, we prune away its pronunciations whose average posterior $\gamma_{wb}$ is too low (e.g. only keeping the top 10), construct a new combined lexicon, and then re-generate the lattices and re-collect acoustic evidence in the same way. We found this pruning is always helpful as it improves the accuracy of the posteriors.
\section{Data-likelihood-reduction based greedy pronunciation selection}
\label{selection}
We formulate the pronunciation selection process as a greedy model selection procedure, with data-likelihood-reduction as the selection criterion. 
In this section, we'll first specify how to compute the optimal data likelihood given a set of pronunciation candidates using EM and propose a pronunciation selection criterion based on likelihood reduction,  and then use an illustrative example to compare the proposed selection criterion against other criteria. 
At last we talk about some practical issues in our algorithm, and summarize the whole iterative framework of pronunciation selection.

\subsection{A pronunciation selection criterion based on  per-utterance likelihood reduction}
Given a set of pronunciation candidates for a specific word $w$, and the conditional likelihood $\tau_{uwb}$ (acoustic evidence) for each utterance $\mathcal{O}_u$, we want to maximize the total data likelihood over the pronunciation model $\boldsymbol{\theta}_w$ \footnote{When we optimize the pronunciation probabilities for a specific word, we consider the pronunciation probabilities for other words as fixed.}:
\begin{eqnarray} \label{eqn:objf}
   \mathcal{L} (\boldsymbol{\theta}_w) = \sum_{u}  \log \left( \sum_b \tau_{uwb} \theta_{wb} \right)
\end{eqnarray}
where the summation $\sum_{u}$ is only taken over utterances where
 the word $w$ actually appears. Since maximizing this objective doesn't have a closed form solution, like \cite{mcgraw2013learning}, we use EM which maximizes the following auxiliary function instead ($n$ stands for the iteration index, $\lambda_{uwb}^{n} \triangleq p(w, b | \mathcal{O}_u, \boldsymbol{\theta_{w}}^n)$ is the pronunciation posterior computed at the $n$th iteration) 
\begin{eqnarray}
Q(\boldsymbol{\theta}_w^{n+1}, \boldsymbol{\theta}_w^{n})
= \sum_{u}  \sum_b \lambda_{uwb}^{n}    \log \theta_{wb}^{n+1}
\end{eqnarray}
Maximizing the above function with the constraint $\sum_{b} \theta_{wb}^{n+1} = 1$ gives the M-step:
\begin{eqnarray}\label{mstep}
   \theta_{wb}^{n+1} \leftarrow 
   \frac{\sum_{u}  \lambda_{uwb}^{n} }{
\sum_{u}  \sum_{b} \lambda_{uwb}^{n} }
\end{eqnarray}
According to Bayes' rule, we compute the updated posteriors $\lambda_{uwb}^{n+1} $ as the following:
\begin{eqnarray}\label{estep}
 \lambda_{uwb}^{n+1}  \leftarrow \frac{\tau_{uwb}\theta_{wb}^{n+1}}{ \sum_b \tau_{uwb} \theta_{wb}^{n+1}} 
\end{eqnarray}
which is the E-step.
By running \eqref{mstep} and \eqref{estep} iteratively until convergence, we can find an optimal pronunciation model $\boldsymbol{\theta}^*_{w}$, and evaluate the optimal log-likelihood \eqref{eqn:objf} $ \mathcal{L} (\boldsymbol{\theta}^*_w) $ (denoted as $\mathcal{L}^*$ for simplicity).
In order to evaluate the importance of a specific pronunciation, say, $b$, we remove $b$ from the pronunciation candidate set $\mathcal{B}$, re-initialize the pronunciation model $\boldsymbol{\theta}^{'}_w$ on top of $\mathcal{B} \backslash b$ , and run EM to optimize \eqref{eqn:objf} with the model $\boldsymbol{\theta}^{'}_w$. Writing the likelihood at convergence after removing $b$ as  $\mathcal{L}_b^*$, we can compute the per-utterance likelihood reduction associated with the pronunciation $b$ as:
\vspace{-0.2em}\[ \overline{\Delta \mathcal{L}}_b \triangleq \frac{\Delta \mathcal{L}_b}{M_w}  =\frac{\mathcal{L}^* - \mathcal{L}^*_b  }{M_w},\] \vspace{-0.2em}
This metric reflects the contribution of each pronunciation to the total data likelihood. With this metric, we can iteratively remove least important pronunciations in a greedy fashion, which is efficient. The complete iterative framework is given in Section~\ref{iterative}.
\vspace{-0.21em}
\subsection{An illustrative example}
\label{example}
Here we show an example to illustrate the advantage of pronunciation selection based on the per-utterance log likelihood reduction $\overline{\Delta \mathcal{L}}_b$  over the learned pronunciation probabilities $\boldsymbol{\theta}^*_w$, in terms of dealing with confusability of pronunciation variants. 

In Table~\ref{tab:word-egs}, we listed the pronunciation candidates, average pronunciation posteriors, learned pronunciation probabilities, and the per-utterance log likelihood reduction of two English words `machine' and `us' taken from the TED-LIUM~\cite{rousseau2012ted} training corpus. Note that the two pronunciations of `machine' only differ in one vowel, while the two pronunciations of `us' represent two distinct meanings. 

We want a selection criterion under which it's possible to put a threshold to rule out the reduction `M IH SH IY N' (generated from phonetic-decoding) in the `machine' case,
while keeping the acronym `Y UW EH S' in the `us' case. Looking at the learned pronunciation probabilities $\boldsymbol{\theta}^{*}_w$, it gives lower values  for  `Y UW EH S' than `M IH SH IY N', and thereby cannot serve as the criterion we need.
However, the per-utterance log likelihood reduction $\overline{\Delta \mathcal{L}}_b$  of `AH S' is much larger than  `M IH SH IY N'  (0.034 v.s. 0.004). Thus it's possible to set a proper threshold on $\overline{\Delta \mathcal{L}}_b$ to keep `AH S' and remove `M IH SH IY N'.

The underlying reason is that the confusability between pronunciations is reflected in the sharpness of the per-utterance pronunciation posteriors $\gamma_{uwb}$. In the `us' case, the two pronunciation variants cannot easily model each other, and therefore the posteriors are very sharp for most examples. Thereby removing the minor pronunciation `Y UW EH S' would result in a greater reduction in the data likelihood. Thus, beyond reflecting the relative frequency, the proposed criterion $\overline{\Delta \mathcal{L}}_b$ is capable of modeling the confusability between pronunciation candidates, which is preferable from the Maximum Likelihood point of view and therefore could help us to select an informative set of pronunciations.

\begin{table} 
\centering
\small
\caption{\label{tab:word-egs} The pronunciation candidate set $\mathcal{B}$, learned pronunciation probabilities $\boldsymbol{\theta}^{*}_w$,  and the per-utterance log likelihood reduction $\overline{\Delta \mathcal{L}}_b$ for two English words `machine' and `us' from TED-LIUM.}
\begin{tabular}{c|c|c}
$w$ & `machine'  & `us' \\\hline
$\mathcal{B}$ & \scriptsize{[`M AH SH IY N', `M IH SH IY N']} &  \scriptsize{[`AH S', `Y UW EH S']} \\
 $\boldsymbol{\theta}^{*}_w$  & [0.987, 0.013] & [0.992, 0.008] \\
  $\overline{\Delta \mathcal{L}}_b$  & [3.575, 0.004] & [15.576, 0.034] \\
\end{tabular}
\end{table}
\subsection{Refining the pronunciation selection criterion $\overline{\Delta \mathcal{L}}_b$}\label{practical}
One difficulty of directly using  $ \overline{\Delta \mathcal{L}}_b$ in an iterative pronunciation selection framework is that, we need to develop an interpretable threshold $T$ in order to decide when to stop removing pronunciations.  However, we notice the upper bound of $ \overline{\Delta \mathcal{L}}_b$ can be achieved in an extreme case, where we remove an absolutely dominating pronunciation $p$ (meaning: the observed conditional likelihoods satisfy: $\tau_{uwp} = 1,\quad \tau_{uwb} = \delta, \quad \forall b \neq p$ ). Before removing $p$, it's obvious from  \eqref{eqn:objf} that the maximum $\mathcal{L}(\boldsymbol{\theta}_w^*) = 0$ can be reached with $\boldsymbol{\theta}_w^*$ being a one-hot vector s.t. $ \theta_{wp} =1$. After removing $p$, with the constraint $\sum_{b \in \mathcal{B}\backslash p} \theta_{wb}^{'}=1$, the log-likelihood is a constant: $\mathcal{L}(\boldsymbol{\theta}_w^{'})  \equiv M_w \log{\delta}$. Then we have: $\overline{\Delta \mathcal{L}}_{p} = (0 - M_w \log{\delta}) / M_w = -\log \delta$.  According to this, we scale this upper bound by a scalar $\alpha$ between $[0,1]$ to get an interpretable threshold: $T= -\alpha \log \delta$ , where $\alpha  = 1$ corresponds to the above extreme case, which means, for a pronunciation to be not removed, it would have to be present with probability 1 in 100$\%$ instances of the word, and  $\alpha  = 0$ means we will never remove any pronunciation candidates. In practice, it's set between $0.005$ and $0.2$. We also make $\alpha$ dependent on the source $s(b)$ of the pronunciation,  which enables us to use a more conservatively threshold for selecting pronunciations from a source where the candidates' quality is lower in general, like phonetic-decoding (pd), e.g. by setting $\alpha_{g2p} = 0.02, \alpha_{pd} = 0.01$. So, we define the ``score" of a pronunciation candidate as ``how far away" its $\overline{\Delta \mathcal{L}}_b$ is to the corresponding threshold, i.e.:
\[ q_b \triangleq  \overline{\Delta \mathcal{L}}_b - T_{s(b)} = \frac{\Delta \mathcal{L}_b}{M_w + \beta_{s(b)}} + \alpha_{s(b)} \log \delta
\]

In our framework we iteratively prune the pronunciation with the lowest score and terminate pruning when all pronunciation have positive scores. Note that the count $M_w$ is smoothed with a source-dependent scalar $\beta_{s(b)}$ (5-15 in practice). The purpose is to keep the score from being to high when $M_w$ is small,  so that in general we select fewer pronunciations if we only have a few acoustic examples of a word.

\subsection{Summary: an iterative framework}
\label{iterative}
The proposed pronunciation selection algorithm, which iteratively prunes pronunciations from the initial candidate set $\mathcal{B}$, is summarized as Algorithm~\ref{alg:iterative} ($\mathcal{B}_t$ stands for the selected subset of pronunciation candidates at iteration $t$).
\vspace{-0.21em}
\begin{algorithm}[htb!]
\centering
\small
\caption{\label{alg:iterative} Greedy pronunciation selection}
\begin{algorithmic}
\STATE set $t = 0$, $\mathcal{B}_0 = \mathcal{B}$. 
\STATE \textbf{While} true:
\STATE \hspace{0.25in} {Initialize}   $\boldsymbol{\theta}_w$ uniformly on $\mathcal{B}_t $.
\STATE  \hspace{0.25in} Run \textbf{EM} on $\mathcal{B}_t$ to get  $\boldsymbol{\theta}_{w}^{*}$ and the optimal log-likelihood $\mathcal{L}^*$.
\STATE \hspace{0.25in}   \textbf{For} $b$ in $\mathcal{B}_t$:
 \STATE \hspace{0.5in}  Initialize  $\boldsymbol{\theta}^{'}_w$ on $\mathcal{B}_t\backslash b$ and run \textbf{EM} to get the optimal log-likelihood  $ \mathcal{L}_{b}^{*}$.
\STATE \hspace{0.5in}  Compute $\Delta \mathcal{L}_b = \mathcal{L}^* -  \mathcal{L}^*_{b}$
\STATE \hspace{0.5in}  Compute $ q_b = \frac{\Delta \mathcal{L}_b}{M_w + \beta_{s(b)}} + \alpha_{s(b)} \log \delta$
\STATE \hspace{0.25in} \textbf{If} $\min\limits_{b \in \mathcal{B}_t} q_b \geq 0$:  
\STATE \hspace{0.5in} \textbf{Output} $\mathcal{B}_t$ as the optimal pronunciation subset.
 \STATE \hspace{0.5in}  \textbf{Break}.
\STATE \hspace{0.25in} \textbf{Else}:
\STATE \hspace{0.5 in}    $\hat{b} = \argmin_{b \in \mathcal{B}_t} q_b$:  
\STATE \hspace{0.25 in} $\mathcal{B}_{t+1} = \mathcal{B}_{t} \backslash \hat{b}$
\STATE \hspace{0.25 in} $t = t+1$
\end{algorithmic}
\label{alg:1}
\end{algorithm}

\section{Experiments}
\label{exp}
In order to evaluate the performance of the proposed lexicon learning framework, a small seed lexicon is built by randomly sampling a small portion ($5\%$) of words from the vocabulary of the expert lexicon of each task. With the seed lexicon, we train a G2P model using Sequitur \cite{bisani2008joint} and apply it to all OOV (w.r.t the seed lexicon) words in the vocabulary of the expert lexicon, to get the "G2P-extended" lexicons.\footnote{In this paper we focus on lexicon learning for alphabetic languages. Thereby a G2P model trained with a small seed lexicon is able to generate pronunciations for most words in the expert lexicon.}  A baseline system called G2P-ext is built using a G2P-extended lexicon with the optimal number of variants per-word tuned on dev data, and another baseline system called G2P-1best is built using a G2P-extended lexicon where we only take the top G2P pronunciation for each word. With this G2P model and acoustic training data for each task, we can build a learned lexicon using the proposed framework, and then train an ASR system called ``Lex-learn".  Besides, we have an ASR system trained using the full expert lexicon as the ``Oracle" system. Note that the training recipes of three ASR systems (G2P-ext, G2P-1best, Oracle, and Lex-learn) for each task only differ in the lexicons (with the same vocabulary). All experiments were done with Kaldi~\cite{povey2011kaldi}.

\begin{table}[h]
\setlength{\tabcolsep}{0.1cm}
\centering
\small
\caption{\label{libri} ASR Performance on Librispeech (WER on the test-clean set (tuned on WER of LF-MMI systems on dev-clean, without 4-gram LM rescoring) with different lexicon conditions (the average \# pronunciations per word for in-vocab words from acoustic training data, are shown in parentheses). The vocab of the full expert lexicon (a subset of CMUDict) has 200K words. }
\vspace{-0.5em}
\begin{tabular}{c||p{40pt}|p{40pt}|p{40pt}|p{40pt}}
\multirow{2}{*}{ } & \multicolumn{4}{c}{WER} \\ \cline{2-5}
& \tabincell{l} {\small{Oracle} \\ \small{(1.08)}}  & \tabincell{l}{\small {G2P-ext} \\ \small{(5.05)}} & \tabincell{l}{\small {G2P-1best} \\ \small{(1)}} & \tabincell{l} {\small{Lex-learn} \\ \small{(1.42)}} \\
\hline
\hline
SAT	 & 11.32 \%  &  13.11 \%  & 14.57 \%  & 11.53	 \%  \\
LF-MMI & 6.44  \%  & 6.76  \%  & 7.15 \%  & 6.64	 \%
\end{tabular}
\end{table}

We conduct experiments on the Librispeech-460 task~\cite{panayotov2015librispeech}. 
For each lexicon condition, we use the 460h training data subset to build speaker-adaptive trained GMM (SAT) models (the same AM training recipe as the "SAT 460" from ~\cite{panayotov2015librispeech}), on top of which we then train sub-sampled time-delay neural
networks (TDNNs) \cite{peddinti2015time} with the lattice-free MMI (LF-MMI) \cite{povey2016purely} criterion. The WERs are shown in Table~\ref{libri}. It can be seen that the learned lexicon performs better than G2P-extended lexicons, and is close to the oracle lexicon. And the LF-MMI systems are much more robust to the lexicon quality than SAT systems, i.e. the G2P-extended and learned lexicons perform closer to the expert lexicon. The learned lexicon closes $88\%$ (SAT)/ $36\%$(LF-MMI) of the WER gap between the G2P-ext system and the oracle system.  Also, looking at the average number of pronunciations per word, the learned lexicon ($1.42$) is much more compact than the G2P-extended lexicon ($5.05$), and is very close to the G2P-1best lexicon ($1$), though it performs much better than the G2P-1best lexicon by a large gap: $20.9\%$ (SAT) / $7.1\%$(LF-MMI) relatively in WER.

In Table \ref{compare}, we compare the proposed framework with more baseline lexicon expansion approaches, on the Librispeech-460 task (WER of SAT systems), with a smaller seed lexicon containing only $1\%(2K)$ randomly sampled words from the same expert lexicon, in order to make the performance gap between different systems more noticeable. ``G2P-ext", as described before, is a baseline built with a G2P-extended lexicon (with a tuned size). ``$pp$-based selection on G2P candidates" means, we first align acoustic training data with a large G2P-extended lexicon containing all G2P generated candidates (up to $10$ candidates per word), and then use max-normalized pronunciation probabilities \cite{chen2016acoustic} to prune those candidates for each OOV word, with a tuned threshold ($0.4$).  The pronunciation candidate pool here is the same as the G2P-ext system (i.e. G2P candidates only). ``$pp$-based  selection on G2P+PD candidates" uses the same lexicon expansion approach as the former one but we also add candidates from phonetic decoding (PD) before selection. Therefore this baseline has the same candidate pool as the proposed framework. The last system
``likelihood-reduction-based selection on G2P+PD candidates" is the proposed framework (i.e. the ``Lex-learn" systems listed before). For fair comparison, under different lexicon conditions, the acoustic models were re-trained on top of the same acoustic model (the one used in the shown G2P-ext system). It can be seen that adding PD candidates to the candidate pool is crucial to the lexicon quality ($0.82\%$ WER improvement), and the proposed pronunciation selection method solely brings $0.18\%$ WER gain and lowers the number of pronunciations per word from $5.43$ to $1.59$.

\begin{table}[h]
\setlength{\tabcolsep}{0.1cm}
\vspace{-0.21em}
\centering
\small
\caption{\label{compare} ASR performance (WER of SAT systems on the test-clean set, without 4-gram LM rescoring) comparison on Librispeech, with different lexicon expansion approaches.}
\vspace{-0.5em}
\begin{tabular}{c||p{32pt}}
Lexicon condition (avg. \#pronunciations per word) & WER  \\
\hline
\hline
G2P-ext	 (6.57) & 13.72	\%  \\
\hline
 \tabincell{l} {$pp$-based selection on \\ G2P candidates (3.77) } &	13.06 \%   \\
 \hline
  \tabincell{l} {$pp$-based selection on \\ G2P+PD candidates (5.43) } &12.24 \% \\
   \hline
  \tabincell{l} {likelihood-reduction-based selection on \\ G2P+PD candidates (1.59)} &	12.06 \%
\end{tabular}
\end{table}

\section{Conclusion and future work}
\label{sec:conclusions}
In this paper, we propose an acoustic-data driven lexicon learning framework using a likelihood-reduction based criterion for selecting pronunciation candidates from multiple sources, i.e. G2P and phonetic decoding. With the proposed criterion, the pronunciation candidates are pruned iteratively in a greedy way, based on the acoustic data likelihood reduction caused by removing each candidate. This approach enables us to construct a compact yet informative lexicon. Experiments on different ASR tasks show that, with the proposed framework, starting with a small expert lexicon (containing $0.88K$ to$10K$ words), we are able to learn a lexicon which performs closer to a full expert lexicon in terms of WER performance on test data, than lexicons built using G2P alone or with a pruning criterion based on pronunciation probabilities. As future work, we'd like to investigate how the amount of training data affects the lexicon learning performance.  
\bibliographystyle{IEEEtran}

\newpage
\bibliography{refs}

\end{document}